\documentclass[letterpaper, 10 pt, conference, final]{ieeeconf}  % Comment this line out if you need a4paper

\IEEEoverridecommandlockouts                              % This command is only needed if 
\usepackage{graphics} % for pdf, bitmapped graphics files

\usepackage[table,xcdraw]{xcolor}
\usepackage{color}
\usepackage{multirow}
\usepackage{array}
\newcolumntype{L}[1]{>{\raggedright\let\newline\\\arraybackslash\hspace{0pt}}m{#1}}
\newcolumntype{C}[1]{>{\centering\let\newline\\\arraybackslash\hspace{0pt}}m{#1}}
\newcolumntype{R}[1]{>{\raggedleft\let\newline\\\arraybackslash\hspace{0pt}}m{#1}}

\newcommand{\etal}{\textit{et al}.}
\usepackage{gensymb}

\usepackage{graphicx}
\usepackage{subfig}
\graphicspath{{./Table_Figures/}}
\title{\LARGE \bf
A Pilot Study on Using an Intelligent Life-like Robot as a Companion for Elderly Individuals with Dementia and Depression
}

\author{Hojjat Abdollahi$^{1}$, Ali Mollahosseini$^{1}$, Josh T. Lane$^{2}$, and Mohammad H. Mahoor$^{1,2}$% <-this % stops a space
\thanks{ \textit{Email addresses:}
	 Hojjat Abdollahi: habdolla@du.edu, 
	Ali Mollahosseini: ali.mollahosseini@du.edu,
    Josh T. Lane: josh.lane@dreamfacetech.com, 
    Mohammad H. Mahoor: mmahoor@du.edu\newline 
}%
\thanks{$^{1}$Department of Electrical and Computer Engineering, University of Denver, CO, USA}%
\thanks{$^{2}$DreamFace Technologies, LLC., Lone Tree, CO, USA
 {\tt\small http://dreamfacetech.com/}}%
}

\begin{document}

\maketitle
\thispagestyle{empty}
\pagestyle{empty}

%%%%%%%%%%%%%%%%%%%%%%%%%%%%%%%%%%%%%%%%%%%%%%%%%%%%%%%%%%%%%%%%%%%%%%%%%%%%%%%%
\begin{abstract}

This paper presents the design, development, methodology, and the results of a pilot study on using an intelligent, emotive and perceptive social robot (aka Companionbot) for improving the quality of life of elderly people with dementia and/or depression. Ryan Companionbot prototyped in this project, is a rear-projected life-like conversational robot. Ryan is equipped with features that can (1) interpret and respond to users' emotions through facial expressions and spoken language, (2) proactively engage in conversations with users, and (3) remind them about their daily life schedules (e.g. taking their medicine on time). Ryan engages users in cognitive games and reminiscence activities. We conducted a pilot study with six elderly individuals with moderate dementia and/or depression living in a senior living facility in Denver. Each individual had 24/7 access to a Ryan in his/her room for a period of 4-6 weeks. Our observations of these individuals, interviews with them and their caregivers, and analyses of their interactions during this period revealed that they established rapport with the robot and greatly valued and enjoyed having a Companionbot in their room.
\end{abstract}

%%%%%%%%%%%%%%%%%%%%%%%%%%%%%%%%%%%%%%%%%%%%%%%%%%%%%%%%%%%%%%%%%%%%%%%%%%%%%%%%
\section{INTRODUCTION}

Developing and studying robots as an assistive tool for health-care professionals is a growing area of research due to the rapid growth in the number of elderly people and the demand for specialized caregivers.
Socially Assistive Robotics (SAR)~\cite{feil2005defining} focus on improving elderly people's quality of life, mental health, and socio-emotional well-being. Social robots are used as companions~\cite{Taggart2005} or therapeutic play partners~\cite{Leite2010}. The essential feature that defines SAR is using social interactions rather than physical interaction to help the user~\cite{Rabbitt2015}. The focus of this paper is on SAR and the companionship it provide for elderly people with moderate depression and/or dementia.

 Dementia is an overall term for diseases that deteriorate individuals' memory and other mental skills. Dementia can significantly reduce elderly individuals' ability to live independently and safely in their homes. It is one of the costliest diseases and requires hours of specialized care-giving for each person~\cite{abbott2011dementia}. Associated to the decline in cognitive abilities, depression is one of the symptoms of dementia~\cite{marti2006socially}.

There is thus a critical and growing demand in the community to find effective ways to provide care for elderly people with dementia. There is an emerging research field in robotics that aims to use social robots to engage effectively in social and conversational interaction with elderly individuals with dementia to improve their socio-emotional behaviors, cognitive functions and well-being. We conducted a pilot study to demonstrate the feasibility of using Ryan Companionbot, a perceptive and empathic conversational humanoid robot, to improve the quality of life of elderly individuals with moderate dementia and/or depression. In this study, we are using spoken dialog combined with a rich list of other stimuli such as eye gaze, head movement, and facial expressions as the primary form of communication between the subject and the robot. Specifically, the objective of this study is to evaluate the following fundamental research questions: 

\begin{enumerate}
	\item \textbf{Long-Term Companionship:} Would enriching the robot with a number of different features keep the subjects engaged over	an extend period of time? 
	
	\item \textbf{Likability and Acceptance:} Is interacting with SAR enjoyable for elderly individuals and do they accept a robot as a companion? 
	
	\item \textbf{Robot Features:} Do the results of the pilot study show that each individual looked for different features  (e.g., spoken dialog system, cognitive games, family photo album narration, music playing, etc.) in the robot?
\end{enumerate}

The remainder of this paper is organized as follows. Section~\ref{sec:RelatedWork} reviews the related work on SAR and employing social robots in elder care. Section~\ref{sec:Ryan} introduces Ryan Companionbot, and explains the software and hardware aspect of Ryan. Section~\ref{sec:Pilot} explains the experiment setting and the methodology of our pilot study to evaluate the above fundamental research questions. Section~\ref{sec:Result} presents the results and analysis of the experiments. The results are categorized in four subsection: long-term companionship, likability and acceptance, caregivers' feedback, and robot features. Finally, Section ~\ref{sec:Conclusion} concludes the paper.

%%%%%%%%%%%%%%%%%%%%%%%%%%%%%%%%%%%%%%%%%%%%%%%%%%%%%%%%%%%%%%%%%%%%%%%%%%%%%%%%
 \section{Related Work}
\label{sec:RelatedWork}

Using SAR to help elderly individuals has recently become more relevant due to the increase in the number of elderly people, the decrease in the cost of technology, and the recent advances in artificial intelligence~\cite{leite2015long}. Residents of nursing homes are living alone with disabilities while in most cases their cognitive abilities are degrading due to old age or various type of dementia~\cite{Kotwal_2016}. Studies suggest that social support for elderly individuals could improve their cognitive function~\cite{zamora2017association}. Using SARs with a focus on the socialization aspect of Human-Robot Interaction (HRI) is a viable option to reduce the burden on caregivers while providing companionship for elderly people, improving their quality of life, and avoiding depression and further degradation of their mental abilities.

 Wada \etal~\cite{wada2003effects} used the robot Paro to study the long-term effect of social robots on residents of a senior care center. The results indicated that elderly residents established a relationship with the robot, developed stronger social ties among themselves, and also maintained a lower stress level. However, Paro lacks the ability to talk and communicate. It is shown that for a social robot to be accepted more easily it should be communicative~\cite{heerink2006studying} and must employ a form of communication with which humans are habituated~\cite{kramer2012human}.

Another key aspect to having a robot as a companion, is continuous (uninterrupted) companionship, meaning having access to the robot at all times.  Autonomy plays a crucial role in achieving an uninterrupted companionship. Most of the studies carried out with social robots on elder care are either done in a Wizard-Of-Oz (WOO) manner~\cite{Vardoulakis2012}, or were limited to a specific scenario~\cite{Pineau_2003}. Vardoulakis \etal~\cite{Vardoulakis2012} designed an experiment to study long-term social companion for older adults. They used a WOO method, and the subject had a robot at his/her home for one week. But since the robot was controlled remotely by an operator, the subject interacted with the robot for only one hour every day. Employing WOO method forces the subjects to use the robot at a specific time of the day for a short period which resembles visiting a friend than having a companion at home. Social robots such as Paro are autonomous and provide continuous companionship, but lack the ability of having a robust social interaction such as spoken dialog and an expressive face.

Deep social interaction is required when dealing with elderly individuals with dementia. Different robots such as Aibo, Paro, and Bandit have been used in studies on the care of elderly people with dementia~\cite{mordoch2013use}. Most of the robots that have been used in these studies have not been built with the social aspect in mind. But to be able to communicate with elderly people with dementia and try to engage them in conversations and games, we need a robot that has been designed to accomplish these social goals. In the following section, we will introduce a robot designed to be social.

%%%%%%%%%%%%%%%%%%%%%%%%%%%%%%%%%%%%%%%%%%%%%%%%%%%%%%%%%%%%%%%%%%%%%%%%%%%%%%%%
\section{Ryan, the Companionbot}
\label{sec:Ryan}
 
The robot used in this study is Ryan Companionbot~\cite{dreamfacetech} which is based on the Expressionbot~\cite{Mollahosseini2014}. Ryan has been developed in DreamFace Technologies, LLC. with the social aspect of HRI in mind. This robot has an emotive and expressive face with accurate visual speech. Ryan can maintain a spoken dialog, recognize expressions on the user's face, and it is equipped with a screen on its torso with features such as cognitive games, music player, narrated photo album, and video player. 

To keep the subject engaged for an extensive period of time, SARs must be personalized~\cite{castellano2008long}. Thus, Ryan was customized for each subject. To increase intimacy and invoke rapport, subjects were allowed to choose the name for the robot. It is worth mentioning that one subject named the robot after his late wife.
Leaving the robot in elderly people's home and having 24/7 access to the robot may cause them to lose motivation. To provoke subjects to act on intrinsic motivation, we had to define tasks and modify Ryan to be enjoyable and not repetitive.

After a while that the human exhausts all of the features of the robot, they will lose interest in interacting with the robot. It is shown that the novelty effect of SARs disappears quickly~\cite{you2006robot}. As the novelty aspect wears off, the social effect might decrease as well~\cite{Fernaeus2010}. By endowing Ryan with a character and a sense of humor on top of various other features implemented into Ryan, we keep the subjects interested to interact with the robot for a long period.

In the next sections, we will explain the hardware and software aspect of the Ryan.

\subsection{Hardware} 
\begin{figure}
	\centering
	\includegraphics[width=0.7\linewidth]{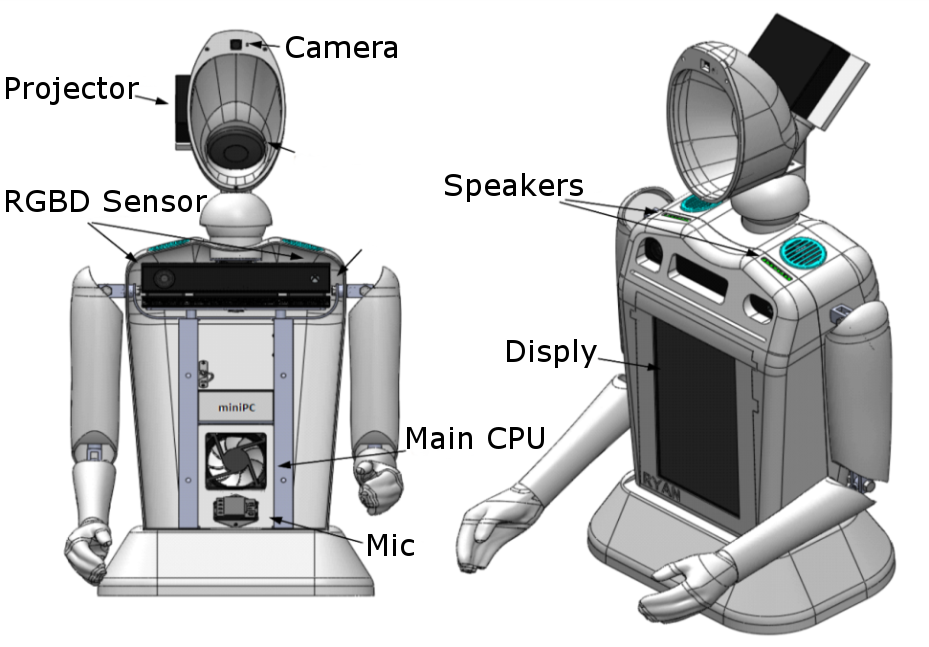}
	\caption{Ryan hardware}
	\label{fig:whole-body}
\end{figure}

Ryan Companionbot hardware is designed with three main components (Figure \ref{fig:whole-body}): 1) the head projection system, 2) the neck mechanism, and 3) the torso.

\subsubsection{Head Projection System} Using a large number of actuators to build a human-like robotic face capable of showing different emotions and visual speech is difficult and expensive~\cite{mollahosseini2014expressionbot}. To avoid tremendous effort required to develop a robotic head capable of having accurate visual speech,  state-of-the-art character animation technology was used to produce an avatar. Using rear projection optics, the head projection system displays the animated avatar onto a mask. This system also allows us to further customize the appearance of the robot. Please consult the work by Mollahosseini \etal~\cite{mollahosseini2014expressionbot} for more details on the projection system.

\subsubsection{Neck Mechanism} The movement of the head for tracking faces and head gestures is controlled by the neck mechanism, a two degree of freedom pan/tilt unit. Having only two degrees of freedom keeps the system simple and suffices for face tracking. The neck has a range of motion of 30\degree of flexion and extension ($\pm$30\degree{} pitch) and 180\degree lateral rotation ($\pm$90\degree yaw). This range allows the head to track the user anywhere in front of the robot. 

\subsubsection{Torso} The main computer, a RGBD camera, a touch screen display, and the power supplies are enclosed inside the torso. Adding a touch screen to the robot added a new way of interacting with Ryan (touch) and also it added the feature to be able to display more information to the user. The display was used for cognitive games, music player, video player, and the narrated photo album. The RGBD camera enables us to have a 3D view of the environment for better tracking the user and also for future studies on activity recognition.

\subsection{Software}
To make Ryan an intelligent and sociable robot that can understand human language and can communicate through spoken dialog, a series of features have been implemented on the robot. Ryan must be able to find the user in the environment, read the user's facial expression, understand user's speech, generate an appropriate response, and say it to the user through audio, accompanied with visual speech while showing a relevant expression on the face. Ryan is also able to communicate with the users through the touch screen on the torso. 

The Microsoft Kinect sensor V2.0~\cite{kinect} acts as the eyes of the system to constantly monitor user's activities and its face detection feature enables Ryan to find the subject in the room. For facial emotion recognition, Ryan uses the Intel RealSense SDK~\cite{realsense} which provides seven basic facial expressions. Intel RealSense SDK is also used as the speech to text engine. Ryan uses the speech emotion recognition Aylien~\cite{aylien} system which is an online natural language processing service for sentiment analysis of the user's speech. A retrieval-based open dialog management systems available on the web (ChatBot\slash Pandorabots~\cite{pandorabots}) is used as the dialog manager.

To reduce subjects' cognitive abilities deterioration, we equipped Ryan with cognitive games focused on patients with dementia. Drugs are not the only method to treat mental diseases such as dementia, Alzheimer's disease, and depression. There exist alternative therapeutic methods such as talking therapies, life story and reminiscence work, and cognitive stimulation therapy for these diseases~\cite{lawrence2012improving}.

 \begin{figure}[!tbp]
 	\centering
 	\subfloat[Main Menu]{\includegraphics[width=0.3\linewidth]{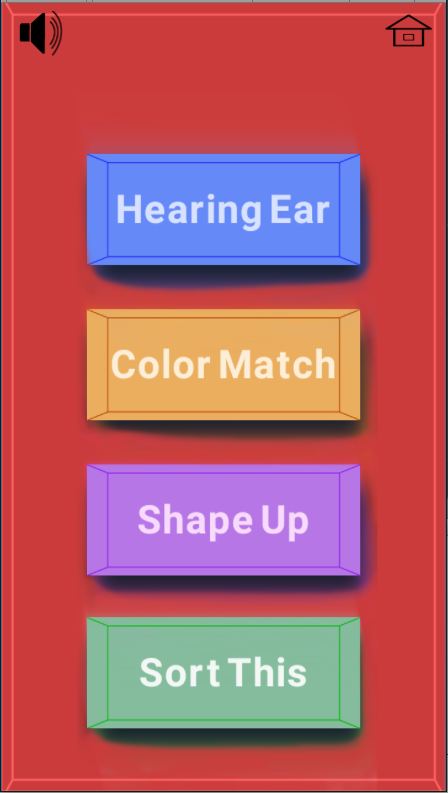}\label{fig:f1}}
 	\hfill
 	\subfloat[Hearing Ear]{\includegraphics[width=0.3\linewidth]{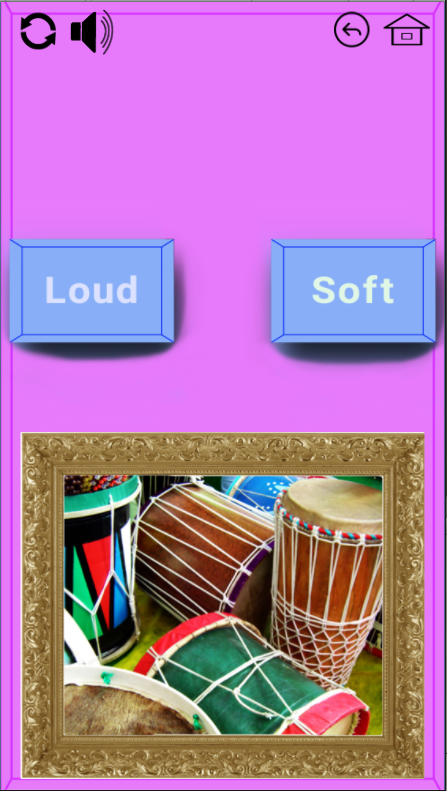}\label{fig:f2}}
 	\hfill
 	\subfloat[Shape Up]{\includegraphics[width=0.3\linewidth]{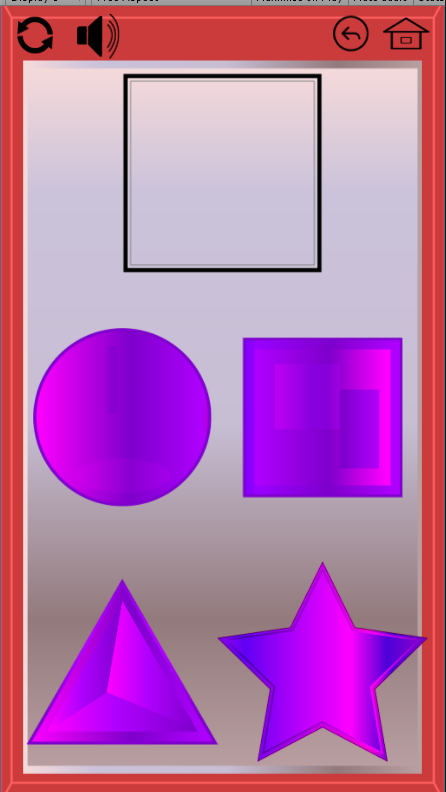}\label{fig:f3}}
 	\caption{Cognitive Games}
 	\label{fig:games}
 \end{figure}

We designed four games (Figure \ref{fig:games}). These games are based on the Montessori-based activities~\cite{judge2000use} to help people suffering from dementia combat the disease. These visual games are simple and interactive with different levels of complexity. The game instructions were given by Ryan and the users could answer the questions either via voice commands or by pushing the buttons on the screen.

There is evidence that life story, photo albums, and reminiscence work, particularly when done one-on-one, can improve mood, well-being and some mental abilities such as memory~\cite{lawrence2012improving}. For each subject we collected about 15-20 old photos and the stories about the event in the photos either from the participant or their close relatives. The photos are shown on the torso screen one-by-one and the robot reads the story back to the user. Sometimes simple questions are asked to engage the user in the conversation.

Reminiscence and memory work also involves talking about things from the past, using prompts such as photos, familiar objects or playing music. A video player application was created to randomly select and play videos from a list of YouTube video clips. The list contained URLs of short (4-5 minutes) YouTube videos queried based on the users' topics of interests (e.g. healthy foods, sports, and nature).

%%%%%%%%%%%%%%%%%%%%%%%%%%%%%%%%%%%%%%%%%%%%%%%%%%%%%%%%%%%%%%%%%%%%%%%%%%%%%%%%
\section{Pilot Study}
\label{sec:Pilot}
\begin{figure}
	\centering
	\includegraphics[width=0.7\linewidth]{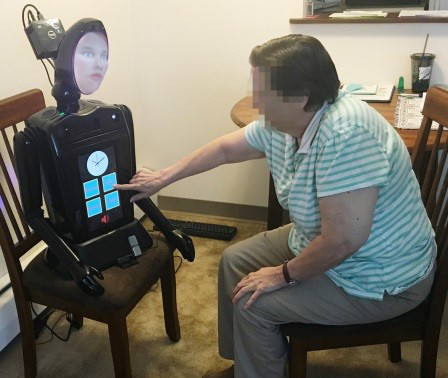}
	\caption{A subject interacting with the robot in her home.}
	\label{fig:interaction21}
\end{figure}

To assess Ryan's feasibility as a Companionbot, we conducted a pilot study with six elderly individuals with dementia and depression living in the Eaton Senior Community in Denver, Colorado~\cite{WinNT}. The robot was left in their home and they had access to the robot at all times. Figure~\ref{fig:interaction21} shows a subject interacting with the robot.

\subsection{Subjects}

A group of six volunteered elderly individuals were selected for this study. The selection criteria included those elderly who live alone, were in the early-mild stage of dementia and may suffer from depression. Other selection criteria included the availability for a period of at least four weeks to house and interact with the robot. Selected subjects were consented prior to participating in the study and family members of the subjects were also informed to insure they are aware of the study. 

The Saint Louis University Mental Status (SLUMS) Examination~\cite{tariq2006saint} and the Patient Health Questionnaire (PHQ-9)~\cite{kroencke2001phq} were completed by each patient and scored by the caregiver prior to the experiment. The SLUMS, developed at the Division of Geriatric Medicine, Saint Louis University School of Medicine, is a favorable screening tool for detecting mild cognitive impairment. The PHQ-9 contains nine questions and is a brief and useful instrument for screening, monitoring, and measuring the severity of depression. The SLUMS scores for people with high school educations are interpreted as follows: 27-30: Normal, 21-26: Mild Neurocognitive Disorder, 1-20: Dementia. The PHQ-9 severity scores are mapped as the following: score 5-9: Minimal Symptoms, score 10-14: Minor depression, score 15-19: Major depression, moderately severe, score$>$20: Major depression, severe. Table~\ref{tab:participants} shows the demographics of the patients that participated in our pilot studies.

\begin{table}[]
	\centering
	\caption{Participants demographics, SLUMS and PHQ-9 Scores. Highlighted cells mean that the symptoms
		(i.e. Dementia and Depression) exist in the patient.}
	\label{tab:participants}
	\begin{tabular}{lccccl}
		\hline
		\multicolumn{1}{|c|}{\textbf{Sbj}} & \multicolumn{1}{c|}{\textbf{Age}} & \multicolumn{1}{c|}{\textbf{Gender}} & \multicolumn{1}{c|}{\textbf{\begin{tabular}[c]{@{}c@{}}SLUMS \\ Score\end{tabular}}} & \multicolumn{1}{c|}{\textbf{\begin{tabular}[c]{@{}c@{}}PHQ-9 \\ Score\end{tabular}}} & \multicolumn{1}{c|}{\textbf{\begin{tabular}[c]{@{}c@{}}Living \\ Resident\end{tabular}}} \\ \hline
		\multicolumn{1}{|l|}{1}            & \multicolumn{1}{c|}{63}           & \multicolumn{1}{c|}{F}               & \multicolumn{1}{c|}{\cellcolor[HTML]{C0C0C0}19}                                      & \multicolumn{1}{c|}{\cellcolor[HTML]{C0C0C0}17}                                                              & \multicolumn{1}{l|}{Independent}                                                         \\ \hline
		\multicolumn{1}{|l|}{2}            & \multicolumn{1}{c|}{86}           & \multicolumn{1}{c|}{M}               & \multicolumn{1}{c|}{\cellcolor[HTML]{C0C0C0}21}                                      & \multicolumn{1}{c|}{1}                                                               & \multicolumn{1}{l|}{Independent}                                                         \\ \hline
		\multicolumn{1}{|l|}{3}            & \multicolumn{1}{c|}{78}           & \multicolumn{1}{c|}{F}               & \multicolumn{1}{c|}{29}                                                              & \multicolumn{1}{c|}{\cellcolor[HTML]{C0C0C0}15}                                      & \multicolumn{1}{l|}{Independent}                                                         \\ \hline
		\multicolumn{1}{|l|}{4}            & \multicolumn{1}{c|}{73}           & \multicolumn{1}{c|}{F}               & \multicolumn{1}{c|}{\cellcolor[HTML]{C0C0C0}17}                                      & \multicolumn{1}{c|}{3}                                                               & \multicolumn{1}{l|}{Assisted}                                                            \\ \hline
		\multicolumn{1}{|l|}{5}            & \multicolumn{1}{c|}{71}           & \multicolumn{1}{c|}{F}               & \multicolumn{1}{c|}{\cellcolor[HTML]{C0C0C0}25}                                      & \multicolumn{1}{c|}{7}                                                               & \multicolumn{1}{l|}{Assisted}                                                            \\ \hline
		\multicolumn{1}{|l|}{6*}            & \multicolumn{1}{c|}{79}           & \multicolumn{1}{c|}{F}               & \multicolumn{1}{c|}{28}                                                              & \multicolumn{1}{c|}{\cellcolor[HTML]{C0C0C0}16}                                      & \multicolumn{1}{l|}{Assisted}                                                            \\ \hline
		\multicolumn{6}{l}{\begin{tabular}[c]{@{}l@{}}\scriptsize{* Subject 6 participated 24 days since she became ill and hospitalized}\\ 
				\scriptsize{ at the end of pilot study} \end{tabular}
		}                                                                                                                                                                                                                                                                                                                                                   
	\end{tabular}
\end{table}

\subsection{Method}

In order to measure how effectively Ryan can provide companionship for elderly individuals with dementia, we conducted a one-on-one (robot vs human) pilot study in the Eaton Senior Community Center. Three Ryan Companionbots were manufactured for the study. Each subject had 24/7 access to Ryan in their rooms for a period of 4-6 weeks. The robot was left in the room of the elderly participant, and he/she treated Ryan Companionbot as his/her guest. To avoid any maintenance issues, the research team monitored the status of the robots remotely. 

Each subject was interviewed to obtain their daily schedules, a set of photos for the album, topics of interest for YouTube video search, and a collection of favorite music and songs. Ryans were customized for each participant. They could call the robot with any name at their preferences. Participants' daily schedule, including reminders to take their medications, were set manually for each subject. 

During the study, all subjects' interactions with Ryan, the facial emotion of the users, the conversations between Ryan and the participants as well as the sentiment of the speech were logged. We analyzed the log files and computed a measurement to evaluate user interactions with Ryan during the pilot study.

%%%%%%%%%%%%%%%%%%%%%%%%%%%%%%%%%%%%%%%%%%%%%%%%%%%%%%%%%%%%%%%%%%%%%%%%%%%%%%%%
\section{Results}
\label{sec:Result}
\subsection{Long-Term Companionship}
In order to measure whether Ryan can be a companion of elderly individuals in long-term, the conversations between Ryan and the participants were recorded over the period of the experiment. The conversations were on different topics such as sports, emotional states, technology, or other topics. Each conversation contains several dialogs between the subjects and Ryan. We defined a dialog as an exchange of one inquiry and response between the subject and Ryan. On average the subjects and Ryan had 198 ($\sigma$=49.2) dialogs per day, with the average length of 9.2 words per each dialog. 

\begin{figure}
	\centering
	\includegraphics[clip, trim=0.5cm 6.5cm 0.5cm 6.5cm,width=\columnwidth]{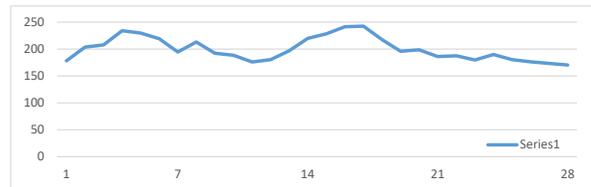}
	\caption{The average number of dialogs between participants and Ryan has not decayed over a period of four weeks (One subject interacted with the robot for three weeks).}
	\label{fig:averageDialogs}
\end{figure}

Figure~\ref{fig:averageDialogs} shows the average number of dialogs of all participants over the period of four weeks. Since SN6 became ill and hospitalized at the end of the pilot study, she only participated 24 days. Therefore, the average shown for the last 4 days are data from 5 subjects. The average number of dialogs time series (shown in Fig.~\ref{fig:averageDialogs}) is then smoothed using a moving average with the window size of five, due to variation between consecutive days and subjects schedule. As shown, the average number of dialogs per day for all subjects did not decay over four weeks. In other words, The subjects kept their interest in having conversations with Ryan even after a long period of time. 
   
The subjects also spent approximately two hours and ten minutes per day interacting with Ryan in different tasks such as playing cognitive games, having conversations, viewing family photo albums, listening to music, etc. Taking into account that although the subjects were living in a senior living facility, where the residents had regular wellness programs and group activities (such as playing games, excessing, occupational and physical therapy), they still were interested in spending time with Ryan, and five of them asked for having Ryan in their room for a more extended time. The result of our pilot study indicated that elderly individuals were interested in having a robot as their companion. They have spent great amount of time with the robot and their interests in speaking with the robot did not decay over time. 

\begin{table*}[!htb]
	\centering
	\caption{The mean rank and questions of the exit survey evaluating users' likability and acceptance of interacting with Ryan and its features (1-strongly disagree, 5-strongly agree)}
	\label{tab:questionnaire}
	\begin{tabular}{l|l|c|c|}
		\cline{2-4}
		& Question                                                          & \begin{tabular}[c]{@{}c@{}}Avg. Score\\ $\pm$ (STD)\end{tabular} & \begin{tabular}[c]{@{}c@{}}Cronbach's \\ alpha\end{tabular} \\ \hline
		\multicolumn{1}{|l|}{\multirow{6}{*}{\begin{tabular}[c]{@{}l@{}}Questions About \\ User Interaction \\ with Ryan\end{tabular}}} & Q1. I enjoyed interacting with the robot.                         & 4.17 $\pm$ 0.75                                                  & \multirow{6}{*}{0.930}                                      \\ \cline{2-3}
		\multicolumn{1}{|l|}{}                                                                                                          & Q2. The conversation with the robot was interesting.              & 4.00 $\pm$ 0.89                                                  &                                                             \\ \cline{2-3}
		\multicolumn{1}{|l|}{}                                                                                                          & Q3. Talking with the robot was like talking to a person.          & 3.00 $\pm$ 1.54                                                  &                                                             \\ \cline{2-3}
		\multicolumn{1}{|l|}{}                                                                                                          & Q4. I feel happier when I had the robot as my company.            & 3.67 $\pm$ 1.03                                                  &                                                             \\ \cline{2-3}
		\multicolumn{1}{|l|}{}                                                                                                          & Q5. I would like to have this robot in my home again.             & 3.33 $\pm$ 1.50                                                  &                                                             \\ \cline{2-3}
		\multicolumn{1}{|l|}{}                                                                                                          & Q6. I feel less depressed after talking to the robot.             & 3.33 $\pm$ 1.36                                                  &                                                             \\ \hline
		\hline
		\multicolumn{1}{|l|}{\multirow{10}{*}{\begin{tabular}[c]{@{}l@{}}Questions About \\ Feature of Ryan\end{tabular}}}              & Q7. I liked the robot's facial expressions.                       & 4.17 $\pm$ 0.75                                                  & \multirow{10}{*}{0.924}                                     \\ \cline{2-3}
		\multicolumn{1}{|l|}{}                                                                                                          & Q8. I liked the robot mirroring my facial expressions.            & 3.50 $\pm$ 1.04                                                  &                                                             \\ \cline{2-3}
		\multicolumn{1}{|l|}{}                                                                                                          & Q9. The robot reminder helped me to be on schedule.               & 4.00 $\pm$ 0.63                                                  &                                                             \\ \cline{2-3}
		\multicolumn{1}{|l|}{}                                                                                                          & Q10. I enjoyed the robot playing my favorite music.               & 4.17 $\pm$ 0.40                                                  &                                                             \\ \cline{2-3}
		\multicolumn{1}{|l|}{}                                                                                                          & Q11. I enjoyed the robot playing videos for me.                   & 3.83 $\pm$ 0.75                                                  &                                                             \\ \cline{2-3}
		\multicolumn{1}{|l|}{}                                                                                                          & Q12. The videos were effective and affected my life style.        & 3.50 $\pm$ 1.51                                                  &                                                             \\ \cline{2-3}
		\multicolumn{1}{|l|}{}                                                                                                          & Q13. I enjoyed playing the games.                                 & 3.33 $\pm$ 1.50                                                  &                                                             \\ \cline{2-3}
		\multicolumn{1}{|l|}{}                                                                                                          & Q14. The games helped me train my brain, though they were simple. & 3.17 $\pm$ 1.32                                                  &                                                             \\ \cline{2-3}
		\multicolumn{1}{|l|}{}                                                                                                          & Q15. The games were challenging.                                  & 2.00 $\pm$ 1.54                                                  &                                                             \\ \cline{2-3}
		\multicolumn{1}{|l|}{}                                                                                                          & Q16. I enjoyed watching my photo album shown by the robot.        & 4.33 $\pm$ 0.81                                                  &                                                             \\ \hline
	\end{tabular}
\end{table*}

\subsection{Likability and Acceptance}

At the end of the study, we asked each participant to complete an exit survey of 16 questions about the experiences they had with Ryan according to the 5-point Likert scale (1-Strongly disagree, 5-Strongly agree). These included six questions about user interaction and companionship of Ryan (i.e., how enjoyable they found interacting and having conversations with the robot), and ten questions about features of Ryan (e.g., ability to show facial expressions, cognitive games, memory photo album, music and video players). 

Table~\ref{tab:questionnaire} shows the exit survey questions and participants' average and standard deviation scores accompanied by Cronbach's Alpha~\cite{cronbach1951coefficient} score for the internal consistency and reliability of each category of questions. 

It can be seen that participants gave strong positive responses (score $>$ 3.5) to most questions on interacting with Ryan, such as ``I enjoyed interacting with the robot'', ``The conversation with the robot was interesting.'' As expected, the participant did not believe that ``talking with the robot was like talking to a person'' with an average score of 3$\pm$ 1.54, however, overall felt happier when they had the robot as their company with an average score of 3.67$\pm$ 1.03.

The survey also indicated that the participants liked the robot's features such as its facial expression (4.17 $\pm$ 0.75), reminder (4.00 $\pm$ 0.63), playing music (4.17 $\pm$ 0.40), playing videos (3.83 $\pm$ 0.75) and watching their photo album (4.33 $\pm$ 0.81) . The games were not challenging enough for the participants with the average score of 2.00 $\pm$ 1.54, but they still found value in playing them, since they ``helped me train my brain.'' The games were designed for elderly in a high level of dementia based on the Montessori-based activities to help people suffering from dementia combat the disease. The authors believe that the games were simple and interactive, but they became boring for the people with early-mild stages of dementia (See Table~\ref{tab:participants} for the SLUM score of the participants).   

In summary, the survey revealed that the subjects liked interacting with Ryan and accepted the robot as a companion although it cannot replace human companionship. They also believed the robot helped them maintain their schedule, improved their mood, and stimulated them mentally. The common sentiment among users after the pilot study was best described by one user's comment, ``She [Ryan] was just enjoyable. We were SAD to see her go.'' The Eaton staff and family members expressed enthusiastic support for the project because it had a consistently positive impact on each of the individuals who interacted with Ryan. For instance, the son of one of the participants said ``[Ryan] has brought color and laughter into my mom's life. She laughs whenever she talks about it!''

\subsection{Caregiver's Feedback}
The users' caregiver, a licensed practical nurse with 20 years of experience, provided feedback on the outcome of the pilot study for each participant. The caregiver closely monitored SN1, SN3 and SN6 who were diagnosed with depression. She confirmed that Ryan elevated the user’s mood. In her words: ``SN6 has been so much happier'', ``SN4 would break out in a big smile when we asked her about her experiences'', and ``You can see the improvement in [SN3's] level of depression after the hip surgery thanks to that sassy roommate [Ryan]''. The caregiver noted that the robot was able to establish a deep connection with the subjects.

\subsection{Robot Features}
\begin{figure}
	\centering
	\includegraphics[width=\columnwidth]{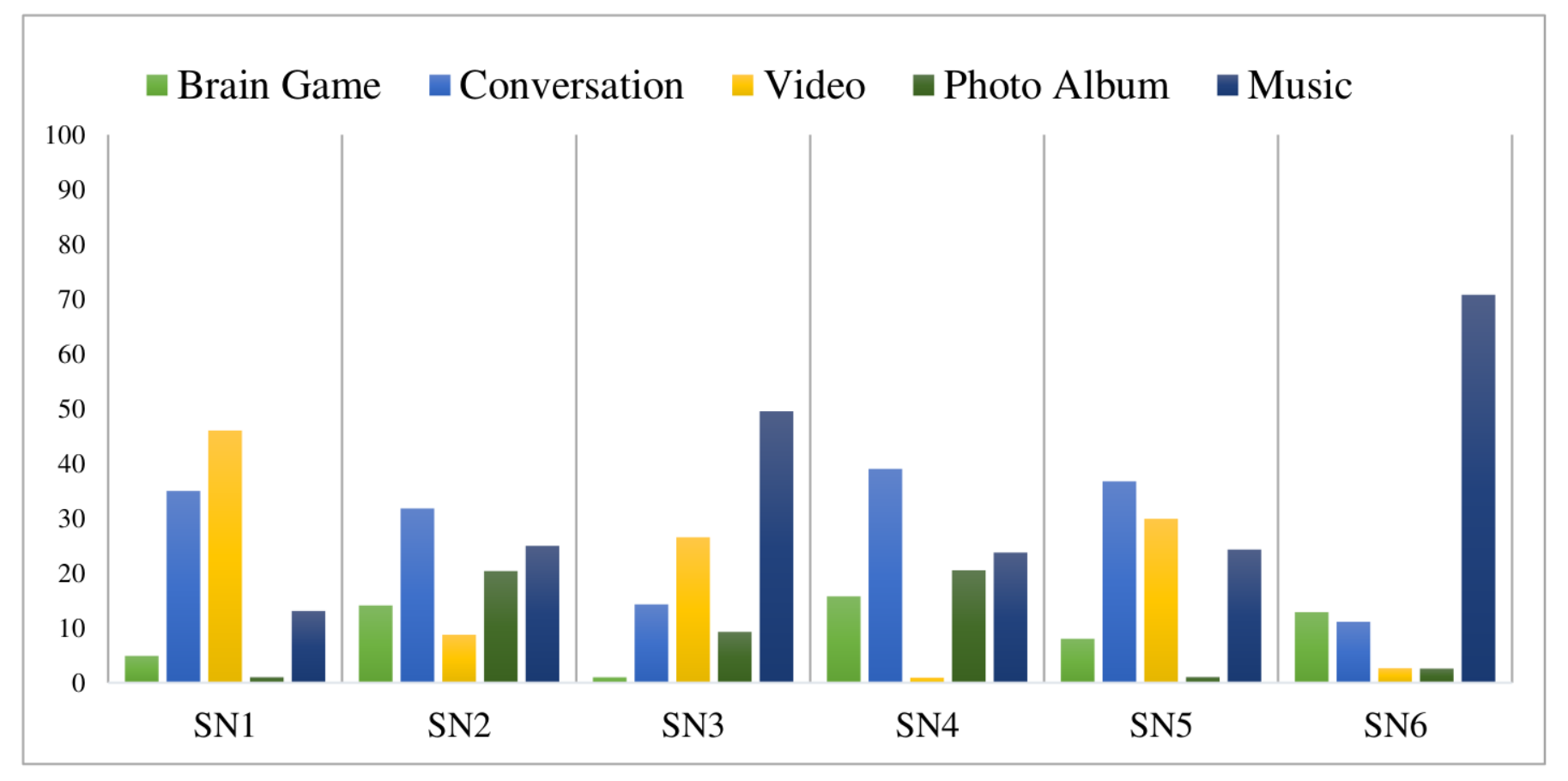}
	\caption{Percentage (\%) of time each user spent in the different activities.}
	\label{fig:averageUsage}
\end{figure}

In order to analyze users' interactions and measure which feature were most appealing for the users, the usage of robot's features were recorded over time. Figure~\ref{fig:averageUsage} shows the percentage of time that each subject spent with different activities (i.e. Games, Conversation, Video, Photo Album, and Music).

As the figure shows, each participant had various interests and found value in different activities, as supported by the users' self-report and caregivers' observations. For example, subjects SN6 and SN3 preferred the music player while SN2, SN4, and SN5 enjoyed the conversation with the robot the most. On average, each user spent approximately two hours and ten minutes per day interacting with Ryan; time that they otherwise would have likely spent alone.
%%%%%%%%%%%%%%%%%%%%%%%%%%%%%%%%%%%%%%%%%%%%%%%%%%%%%%%%%%%%%%%%%%%%%%%%%%%%%%%%%%
\section{Conclusion}
\label{sec:Conclusion}

This paper presented the design, development, and successful integration of a Companionbot to improve the quality of life of elderly individuals with dementia and depression. Three fundamental research questions were posed and addressed in this paper: 1) \textbf{Long-Term Companionship:} Would enriching the robot with a number of different	features keep the subjects engaged over	an extend period of time? 2) \textbf{Likability and Acceptance:} Would elderly individuals accept a robot as a companion? Is interaction with the robot enjoyable to them? 3) \textbf{Robot Features:} Do the results of the pilot study show that each individual looked for different features in the robot? Our experimental results and analysis of the collected data indicated that elderly individuals were interested in having a robot as their companion and their interest did not decay over time.  The subjects liked interacting with Ryan and accepted the robot as a companion although it cannot replace human companionship. The proposed emotionally intelligent conversational Companionbot with a variety of engaging activities can fully engage users and be a promising tool to improve the quality of life of elderly individuals with dementia and depression.

\section{Acknowledgment}
\label{sec:Ack}
This research is partially supported by grants IIP-1548956 and CNS-1427872 from the National Science Foundation.

%%%%%%%%%%%%%%%%%%%%%%%%%%%%%%%%%%%%%%%%%%%%%%%%%%%%%%%%%%%%%%%%%%%%%%%%%%%%%%%%

\bibliographystyle{IEEEtran}
% argument is your BibTeX string definitions and bibliography database(s)
%\bibliography{IEEEabrv,library}

\end{document}